\documentclass[runningheads]{llncs}
\usepackage{graphicx}

\usepackage{tikz}
\usepackage{comment}
\usepackage{amsmath,amssymb} 
\usepackage{bbm}
\usepackage{xspace}
\usepackage{makecell, verbatim, sidecap}
\usepackage{color}
\usepackage{hyperref}
\usepackage{cite}
\usepackage{marvosym}

\usepackage[accsupp]{axessibility}  

\begin{document}
\pagestyle{headings}
\mainmatter
\def\ECCVSubNumber{1182}  

\title{Box-supervised Instance Segmentation with Level Set Evolution} 

\titlerunning{Box-supervised Instance Segmentation with Level Set Evolution}

\author{
	Wentong Li\inst{1}    \and
	Wenyu Liu\inst{1}  \and
	Jianke Zhu\inst{1}\textsuperscript{(\Letter)}        \and  
	Miaomiao Cui\inst{2}  \and \\
	Xiansheng Hua\inst{2}  \and
	Lei Zhang\inst{3}
}

\authorrunning{Wentong Li et al.}

\institute{$^1$Zhejiang University~
	$^2$Alibaba Damo Academy \\
    $^3$The Hong Kong Polytechnic University \\
	\email{{\{liwentong,~liuwenyu.lwy,~jkzhu\}@zju.edu.cn,~miaomiao.cmm@alibaba-inc.com,\\ xshua@outlook.com, cslzhang@comp.polyu.edu.hk}}
}

\maketitle

\begin{abstract}
In contrast to the fully supervised methods using pixel-wise mask labels, box-supervised instance segmentation takes advantage of the simple box annotations, which has recently attracted a lot of research attentions. In this paper, we propose a novel single-shot box-supervised instance segmentation approach, which integrates the classical level set model with deep neural network delicately. Specifically, our proposed method iteratively learns a series of level sets through a continuous Chan-Vese energy-based function in an end-to-end fashion. A simple mask supervised SOLOv2 model is adapted to predict the instance-aware mask map as the level set for each instance. Both the input image and its deep features are employed as the input data to evolve the level set curves, where a box projection function is employed to obtain the initial boundary. By minimizing the fully differentiable energy function, the level set for each instance is iteratively optimized within its corresponding bounding box annotation. 
The experimental results on four challenging benchmarks demonstrate the leading performance of our proposed approach to robust instance segmentation in various scenarios.
The code is available at: \href{https://github.com/LiWentomng/boxlevelset}{https://github.com/LiWentomng/boxlevelset}.

\end{abstract}

\section{Introduction}
Instance segmentation aims to obtain the pixel-wise labels of the interested object, which plays an important role in many applications, such as autonomous driving and robotic manipulation. Though having achieved promising performance, most of the existing instance segmentation approaches~\cite{eccv2020boundary,cvpr2019mask,wang2020solov2,eccv2020_condinst,iccv2017maskrcnn} are trained in a supervised manner, which heavily depend on the pixel-wise mask annotations and incur expensive labeling costs.

To deal with this problem, box-supervised instance segmentation takes advantage of the simple box annotation rather than the pixel-wise mask labels, which has recently attracted a lot of research attentions~\cite{eccv2020box2seg,nips2019-bbtp,cvpr2021_boxinst,cvpr2021bbam,cvpr2021boxcaseg,iccv2021discobox}. To enable pixel-wise supervision with box annotation, some methods~\cite{cvpr2021bbam,cvpr2021boxcaseg} focus on 
generating the pseudo mask labels by an independent network, which needs to employ extra auxiliary salient data~\cite{cvpr2021boxcaseg} or post-processing methods like MCG~\cite{tpmai2017mcg} and CRF~\cite{krahenbuhl2011efficient} to obtain precise pseudo labels. 
Due to the involved multiple separate steps, the training pipeline becomes complicated with many hyper-parameters. Several recent approaches~\cite{nips2019-bbtp,cvpr2021_boxinst} suggest a unified framework using the pairwise affinity modeling, e.g., neighbouring pixel pairs~\cite{nips2019-bbtp} and colour pairs~\cite{cvpr2021_boxinst}, enabling an end-to-end training of the instance segmentation network. The pairwise affinity relationship is defined on the set of partial or all neighbouring pixel pairs, which oversimplifies the assumption that the pixel or colour pairs are encouraged to share the same label. The noisy contexts from the objects and background with similar appearance
are inevitably absorbed, leading to inferior instance segmentation performance.

In this paper, we propose a novel single-shot box-supervised instance segmentation approach to address the above limitations. Our approach integrates the classical level set model~\cite{osher1988fronts,tip2001_active_contour} with deep neural network delicately. Unlike the existing box-supervised methods~\cite{nips2019-bbtp,cvpr2021_boxinst,cvpr2021bbam,iccv2021discobox}, we iteratively learn a series of level set functions for implicit curve evolution within the annotated bounding box in an end-to-end fashion. 
Different from fully-supervised level set-based methods~\cite{CVPR2017hudeep,cvpr2019deeplevelsetevolution,eccv2020levelset,yuan2020deep}, 
our proposed approach is able to train the level set functions in a weakly supervised manner  using only the bounding box annotations, which are originally used for object detection. 

Specifically, we introduce an energy function based on the classical continuous Chan-Vese energy functional~\cite{tip2001_active_contour}, and make use of a simple and effective mask supervised method, i.e., SOLOv2~\cite{wang2020solov2}, to predict the instance-aware mask map as the level set for each instance. In addition to the input image, the deep structural features with long-range dependencies are introduced to robustly evolve the level set curves towards the object's boundary, which is initialized by a box projection function at each step.
By minimizing the fully differentiable energy function, the level set for each instance is iteratively optimized within its corresponding bounding box annotation. 
Extensive experiments are conducted on four challenging benchmarks for instance segmentation under various scenarios, including general scene, remote sensing and medical images. The leading qualitative and quantitative results demonstrate the effectiveness of our proposed method. Especially, on remote sensing and medical images, our method outperforms the state-of-the-art methods by a large margin.

The highlights of this work are summarized as follows:

1) We propose a novel level set evolution-based approach to instance segmentation. To the best of our knowledge, this is the first deep level set-based method that tackles the problem of box-supervised instance segmentation. 

2) We incorporate the deep structural features with the low-level image to achieve robust level set evolution within bounding box region, where a box projection function is employed for  level set initialization.

3) Our proposed method achieves new state-of-the-arts of box-supervised instance segmentation on COCO~\cite{lin2014microsoft} and Pascal VOC~\cite{pascalvoc2010} datasets, remote sensing dataset iSAID~\cite{cvpr2019isaid} and medical dataset LiTS~\cite{bilic2019lits}.

\section{Related Work}
\subsection{Box-supervised Instance Segmentation}
The existing instance segmentation methods can be roughly divided into two categories. The first group~\cite{iccv2017maskrcnn,cvpr2020pointrend,eccv2020boundary,zhang2021refinemask} performs segmentation on the regions extracted from the detection results. Another category~\cite{iccv2019yolact,bolya2020yolact++, eccv2020_condinst,cvpr_2020polarmask,wang2020solov2} directly segments each instance in a fully convolutional manner without resorting to the detection results. However, all these methods rely on the expensive pixel-wise mask annotations. 

Box-supervised instance segmentation, which only employs the bounding box annotations to obtain pixel-level mask prediction, has recently been receiving increasing attention. Khoreva~\textit{et al.}~\cite{cvpr2017SDI} proposed to predict the mask with box annotations under the deep learning framework, which heavily depends on the region proposals generated by the unsupervised segmentation methods like GrabCut~\cite{TOG2004grabcut} and MCG~\cite{tpmai2017mcg}. Based on Mask R-CNN~\cite{iccv2017maskrcnn}, Hsu~\textit{et al.}~\cite{nips2019-bbtp} formulated the box-supervised instance segmentation into a multiple instance learning (MIL) problem by making use of the neighbouring pixel-pairwise affinity regularization. BoxInst~\cite{cvpr2021_boxinst} uses the color-pairwise affinity with box constraint under an efficient RoI-free CondInst framework~\cite{eccv2020_condinst}. Despite the promising performance, the pairwise affinity relationship is built on either partial or all neighbouring pixel pairs with the oversimplified assumption that spatial pixel or color pairs are encouraged to share the same label. This inevitably introduces noises, especially from the nearby background or similar objects. Besides, the recent methods like BBAM~\cite{cvpr2021bbam} and DiscoBox~\cite{iccv2021discobox} focus on the generation of proxy mask labels, which often require multiple training stages or networks to achieve promising performance. Unlike the above methods, our proposed level set-based approach is learned implicitly in an end-to-end manner, which is able to iteratively align the instance boundaries by optimizing the energy function within the box region.

\subsection{Level Set-based Segmentation}
As a classical variational approach, the level set methods~\cite{levelset1995a, osher1988fronts} have been widely used in image segmentation, which can be categorized into two major groups: region-based methods~\cite{tip2001_active_contour, ijcv2002multiphase, mumford1989optimal} and edge-based methods~\cite{ijcv1997geodesic, tpami1995shape}. The key idea of level set is to represent the implicit curve by an energy function in a higher dimension, which is iteratively optimized by using gradient descent. 
Some works~\cite{eccv2020levelset,tip2019mumford, CVPR2017hudeep,cvpr2019deeplevelsetevolution,yuan2020deep} have been proposed to embed the level set into the deep network in an end-to-end manner and achieve promising segmentation results. 
Wang~\textit{et al.}~\cite{cvpr2019deeplevelsetevolution} predicted the evolution parameters and evolved the predicted contour by incorporating the user clicks on the boundary points. The energy function is based on the edge-based level set method in ~\cite{ijcv1997geodesic}. Levelset R-CNN~\cite{eccv2020levelset} performs the Chan-Vese level set evolution with the deep features based on Mask R-CNN~\cite{iccv2017maskrcnn}, where the original image is not used in the optimization.
Yuan~\textit{et al.}~\cite{yuan2020deep} built a piecewise-constant function to parse each constant sub-region corresponding to a different instance based on the Mumford–Shah model~\cite{mumford1989optimal}, which achieves instance segmentation by a fully convolutional network. The above methods perform level set evolution between deep features and ground-truth mask in a fully supervised manner, which train the network to predict different sub-regions and get object boundaries. Our proposed approach performs level set evolution only using the box-based annotations without the pixel-wise mask supervision. 

Kim~\textit{et al.}~\cite{tip2019mumford} performed level set evolution in an unsupervised manner, which is mostly related to our proposed approach. To achieve $N$-class semantic segmentation, it employs the global multi-phase Mumford–Shah function~\cite{mumford1989optimal} that only evolves on the low-level features of the input image. Our method is based on the Chan-Vese functional~\cite{tip2001_active_contour}, which is constrained within the local bounding box with the enriched information from both input image and high-level deep features. Moreover, the initialization of level set is generated automatically for robust curve evolution.

\begin{figure}[t]
	\centering
	\includegraphics[width=1.0\linewidth]{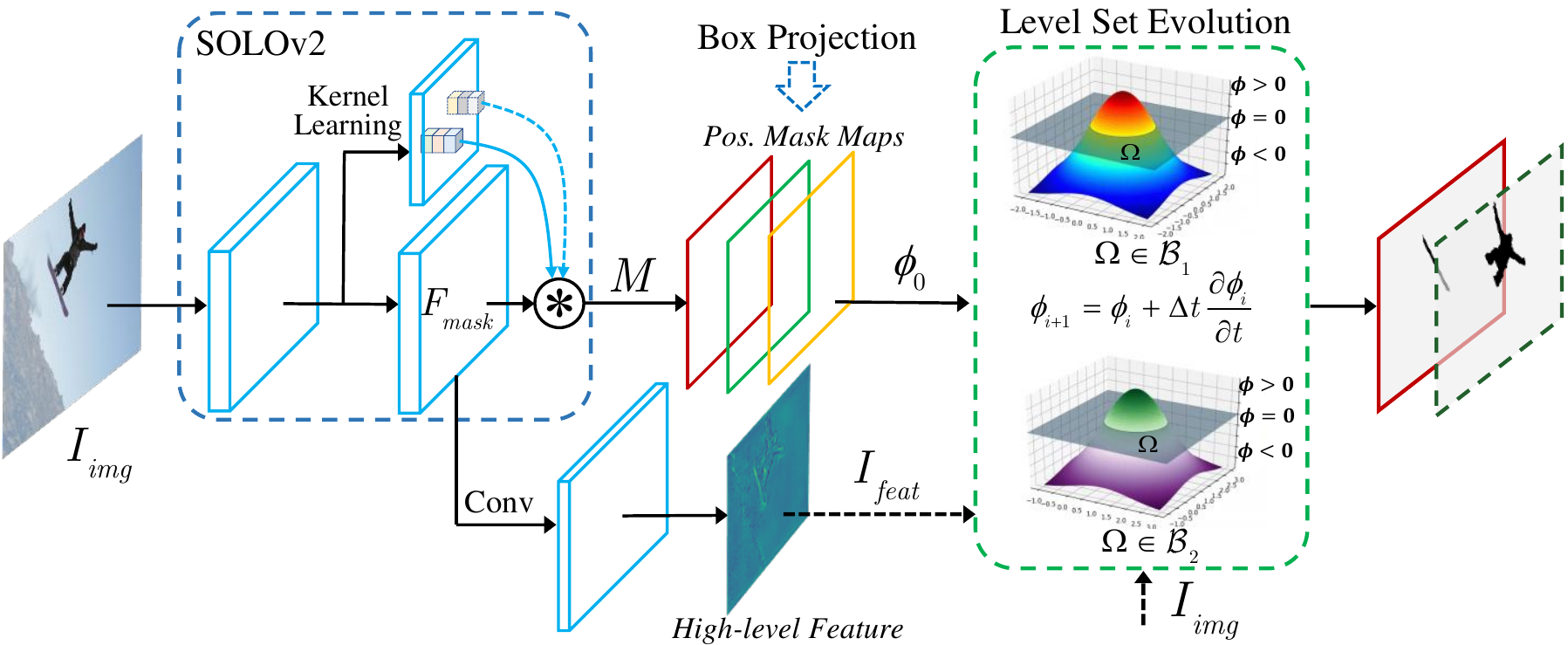}
	\caption{\textbf{Overview of our method.} Our framework is designed based on SOLOv2~\cite{wang2020solov2}. The positive mask maps $M$ are obtained by level set evolution within the bounding box region. 
	With the iterative energy minimization, the accurate instance segmentation can be obtained with box annotations only. The category branch is not shown here for simpler illustration. }
	\label{fig:overallnetwork}
\end{figure}

\section{Proposed Method}
In this section, we present a novel box-supervised instance segmentation method, which incorporates the classical continuous Chan-Vese energy-based level set model~\cite{tip2001_active_contour} into deep neural network. 
To this end, we introduce an energy function, which enables the neural network to learn a series of level set functions evolving to the instance boundaries implicitly. In specific, we take advantage of an effective mask-supervised SOLOv2 model ~\cite{wang2020solov2} to dynamically segment objects by locations and predict the instance-aware mask map of full-image size.
To facilitate the box-supervised instance segmentation, we treat each mask map as the level set function $\phi$ for its corresponding object.
Furthermore, we make use of both the input image $I_{img}$ and high-level deep features $I_{feat}$ as the input to evolve the level set, where a box projection function is employed to encourage the network to automatically estimate an initial level set $\phi_0$ at each step. The level set for each instance is iteratively optimized within its corresponding bounding box annotation. Fig.~\ref{fig:overallnetwork} gives the overview of our proposed framework. 

\subsection{Level Set Model in Image Segmentation}
We first give a brief review of the level set methods~\cite{mumford1989optimal, tip2001_active_contour, ijcv2002multiphase}, which formulate the image segmentation as a consecutive energy minimization problem.
In the Mumford-Shah level set model~\cite{mumford1989optimal}, the segmentation of a given image $I$ is obtained by finding a parametric contour $C$, which partition the image plane $\Omega \subset \mathbb{R}^2$ into $N$ disjoint regions ${\Omega _1}, \cdots ,{\Omega _N}$. The Mumford-Shah energy functional $\mathcal{F}^{MS}(u ,C)$ can be defined as below:
\begin{footnotesize}
\begin{equation} \label{eq2}
\begin{aligned}
{\mathcal{F}^{MS}}({u_1}, \cdots ,{u_N},{\Omega _1}, \cdots ,{\Omega _N}) 
= \sum\limits_{i = 1}^N {(\int\limits_{{\Omega _i}} {{{(I - {u_i})}^2}} dxdy + \mu \int\limits_{{\Omega _i}} {{{\left| {\nabla {u_i}} \right|}^2}dxdy + \gamma \left| {{C_i}} \right|} ),}
\end{aligned}
\end{equation}\end{footnotesize}

\noindent
where $u_i$ is a piecewise smooth function approximating the input $I$, ensuring the smoothness inside each region ${\Omega _i}$. $\mu$ and $\gamma$ are weighted parameters.

Chan and Vese~\cite{tip2001_active_contour} later simplified the Mumford-Shah functional as a variational level set, which has been explored aplenty~\cite{wang2010efficient,liu2012local,mavska2013segmentation,xu2011image}. Specially, it can be derived as follows,
\begin{footnotesize}
\begin{equation} \label{eq3}
\begin{aligned}
{\mathcal{F}^{CV}}(\phi ,{c_1},{c_2}) &= \int\limits_\Omega  {{{\left| {I(x,y) - {c_1}} \right|}^2}H(\phi (x,y))} dxdy \\ & + \int\limits_\Omega  {{{\left| {I(x,y) - {c_2}} \right|}^2}(1 - H(\phi (x,y)))} dxdy + \gamma \int\limits_\Omega  {\left| {\nabla H(\phi (x,y))} \right|dxdy}
\end{aligned}
\end{equation}
\end{footnotesize}

\noindent
where $H$ is the Heaviside function and $\phi(x,y)$ is the level set function, whose zero crossing contour $C = \{ (x,y):\phi (x,y) = 0\}$ divides the image space $\Omega$ into two disjoint regions, 
inside contour $C$: ${\Omega _1} = \{ (x,y):\phi (x,y) > 0\}$ and outside contour $C$: ${\Omega _2} = \{ (x,y):\phi (x,y) < 0\}$. 
In Eq.~(\ref{eq3}), the first two terms intend to fit the data, and the third term regularizes the zero level contour with a non-negative parameter $\gamma$. $c_1$ and $c_2$ are the mean values of input $I(x,y)$ inside $C$ and outside $C$, respectively. The image segmentation is achieved by finding a level set function $\phi(x,y)= 0$ with ${c_1}$ and ${c_2}$ that minimize the energy $\mathcal{F}^{CV}$.

\subsection{Box-supervised Instance Segmentation}
Our proposed method exploits the level set evolution with Chan-Vese energy-based model~\cite{tip2001_active_contour} to achieve high-quality instance segmentation using the box annotations only.

\noindent \textbf{Level Set Evolution within Bounding Box.}  Given an input image $I(x,y)$, we aim to predict the object boundary curve by evolving a level set implicitly within the region of annotated bounding box $\mathcal{B}$.
The mask prediction $M \in {\mathbb{R}^{H \times W \times {S^2}}}$ by SOLOv2 contains ${S \times S }$ potential instance maps of size $H \times W$. Each potential instance map contains only one instance whose center is at location $(i,j)$.
The mask map predicted for the location $(i,j)$ with the category probability $p_{i,j}^* > 0$ is regarded as the positive instance sample. We treat each positive mask map within box $\mathcal{B}$ as the level set $\phi(x,y)$, and its corresponding pixel space of input image $I(x,y)$ is referred as $\Omega$, i.e., $\Omega  \in \mathcal{B}$. $C$ is the segmentation boundary with zero level $C = \{ (x,y):\phi (x,y) = 0\}$, which partitions the box region into two disjoint regions, i.e., foreground object and background. 

To obtain the accurate boundary for each instance, we learn a series of level sets $\phi(x,y)$ by minimizing the following energy function:
\begin{small} 
\begin{equation} \label{eq4}
\begin{aligned}
\mathcal{F}(\phi ,I, {c_1},{c_2},\mathcal{B}) &=  \int\limits_{\Omega  \in \mathcal{B}} {{{\left| {{I^*}(x,y) - c_1} \right|}^2}\sigma (\phi (x,y))} dxdy \\ & +  \int\limits_{\Omega  \in \mathcal{B}} {{{\left| {{I^*}(x,y) - c_2} \right|}^2}(1 - \sigma (\phi (x,y)))} dxdy  + \gamma \int\limits_{\Omega  \in \mathcal{B}} {\left| {\nabla \sigma (\phi (x,y))} \right|dxdy,} 
\end{aligned}
\end{equation}
\end{small}

\noindent
where $I^*(x,y)$ denotes the normalized input image $I(x,y)$, 
$\gamma$ is a non-negative weight, and
$\sigma$ denotes the $sigmoid$ function that is treated as the characteristic function for level set $\phi(x,y)$.  Different from the traditional Heaviside function~\cite{tip2001_active_contour}, the $sigmoid$ function is much smoother, which can better express the characteristics of the predicted instance and improve the convergence of level set evolution during the training process. The first two items in Eq.~(\ref{eq4}) force the predicted $\phi(x,y)$
to be uniform both inside region $\Omega$ and outside area
$\bar{\Omega}$. $c_1$ and $c_2$ are the mean values of $\Omega$ and $\bar{\Omega}$, which are defined as below:
\begin{small}
\begin{equation} \label{eq5}
\begin{aligned}
c_1(\phi ) = \frac{{\int\limits_{\Omega  \in \mathcal{B}} {{I^*}(x,y)\sigma (\phi (x,y))} dxdy}}{{\int\limits_{\Omega  \in \mathcal{B}} {\sigma (\phi (x,y))} dxdy}}, \ \ 
c_2(\phi ) = \frac{{\int\limits_{\Omega  \in \mathcal{B}} {{I^*}(x,y)(1 - \sigma (\phi (x,y)))} dxdy}}{{\int\limits_{\Omega  \in \mathcal{B}} {(1 - \sigma (\phi (x,y)))} dxdy}}.
\end{aligned}
\end{equation}
\end{small}

The energy function $\mathcal{F}$ can be optimized with  gradient back-propagation during training. With the time step $t \geqslant 0$, the derivative of energy function $\mathcal{F}$ upon $\phi$ can be written as follows:
\begin{small}
\begin{equation} \label{eq7}
\begin{aligned}
\frac{{\partial \phi }}{{\partial t}} =  - \frac{{\partial \mathcal{F}}}{{\partial \phi }} =  - \nabla \sigma (\phi )[ {({I^*}(x,y) - c_1)^2} - {({I^*}(x,y) - c_2)^2} + \gamma div\left (\frac{{\nabla \phi }}{{\left| {\nabla \phi } \right|}}\right)],
\end{aligned}
\end{equation}
\end{small}

\noindent
where $\nabla$ and $div$ are the spatial derivative and  divergence operator, respectively. Therefore, the update of $\phi$ is computed by
\begin{equation} \label{eq9}
\begin{aligned}
{\phi _i} = {\phi _{i - 1}} + \Delta t\frac{{\partial \phi_{i-1} }}{{\partial t}}.
\end{aligned}
\end{equation}
The minimization of the above terms can be viewed as an implicit curve evolution along the descent of energy function. The optimal boundary $C$ of the instance is obtained by minimizing the energy $\mathcal{F}$ via iteratively fitting $\phi _i$ as follows:
\begin{equation} \label{eq10}
\begin{aligned}
\mathop {\inf }\limits_{\Omega \in \mathcal{B}} \{ \mathcal{F}(\phi)\}  \approx 0 \approx \mathcal{F}({\phi_{i}}).
\end{aligned}
\end{equation}

\noindent \textbf{Input Data Terms.} The energy function in Eq.~(\ref{eq4}) encourages the curve evolution based on the uniformity of regions inside and outside the object. The input image $I_u$ represents the essential low-level features, including shape, colour, image intensities, etc. 
However, such low-level features usually vary with illumination variations, different materials and motion blur, making the level set evolution less robust.

In addition to the normalized input image, we take into account the high-level deep features $I_f$, which embed the image semantic information, to obtain more robust results. To this end, we make full use of the unified and high-resolution mask feature $F_{mask}$ from all FPN levels in SOLOv2, which is further fed into a convolution layer to extract the high-level features $I_f$. Besides, the features $I_f$ are enhanced by the tree filter~\cite{nips2019learnable,liang2022tree}, which employs minimal spanning tree to model long-range dependencies and preserve the object structure.
The overall energy function for level set  evolution can be formulated as follows:
\begin{equation} \label{eq8}
\begin{aligned}
\mathcal{F(\phi)} = \lambda_1 *  {\mathcal{F}}(\phi ,I_u, c_{{u_1}},c_{{u_2}},\mathcal{B}) + \lambda_2 *  {\mathcal{F}}(\phi, I_f,c_{{f_1}},c_{{f_2}},\mathcal{B}),
\end{aligned}
\end{equation}
where $\lambda_1$ and $\lambda_2$ are weights to balance the two kinds of features. $c_{{u_1}}$, $c_{{u_2}}$ and $c_{{f_1}}$, $c_{{f_2}}$ are the mean values for input terms $I_u$ and $I_f$, respectively.

\noindent \textbf{Level Set Initialization.} Conventional level set methods are sensitive to the initialization that is usually manually labeled. In this work, we employ a box projection function~\cite{cvpr2021_boxinst} to encourage the model to automatically generate a rough estimation of the initial level set $\phi_0$ at each step.

In particular, we utilize the coordinate projection of ground-truth box to $x$-axis and $y$-axis and calculate the projection difference between the predicted mask map and the ground-truth box. 
Such a simple scheme limits the predicted initialization boundary within the bounding box, providing a good initial state for curve evolution. Let $m^b \in {\{ 0,1\} ^{H \times W}}{\rm{ }}$ denote the binary region by assigning one to the locations in the ground-truth box, and zero otherwise. The mask score predictions ${m^p} \in {(0,1)^{H \times W}}$ for each instance can be regarded as the foreground probabilities. The box projection function ${\mathcal{F}(\phi_0)_{box}}$ is defined as below:
\begin{equation} \label{eqbox}
\begin{aligned}
{\mathcal{F}(\phi_0)_{box}} = \mathcal{P}_{dice}({m^p_x},{m^b_x}) + \mathcal{P}_{dice}({m^p_y},{m^b_y}),
\end{aligned}
\end{equation}
where ${m^p_x}$, $m^b_x$ and ${m^p_y}$, $m^b_y$ denote the $x$-axis projection and $y$-axis projection for mask prediction $m^p$ and binary ground-truth region $m^b$, respectively. $\mathcal{P}_{{dice}}$ represents the projection operation measured by 1-D dice coefficient~\cite{ic3dv2016_dice_loss}.  

\subsection{Training and Inference}
\textbf{Loss Function.} The loss function $L$ to train our proposed network consists of two items, including 
$L_{cate}$ for category classification and $L_{inst}$ for instance segmentation with box annotations: 
\begin{equation} \label{loss}
\begin{aligned}
L = {L_{cate}} + {L_{inst}},
\end{aligned}
\end{equation}
where $L_{cate}$ is the Focal Loss~\cite{lin2017focal}. For $L_{inst}$, we employ the presented differentiable level set energy as the optimization objective: 
\begin{equation} \label{levelsetloss}
\begin{aligned}
{L_{inst}} = \frac{1}{{{N_{pos}}}}\sum\limits_k {{\mathbbm{1}_{\{ p_{i,j}^* > 0\} }}\{ }  \mathcal{F}(\phi ) +  \alpha \mathcal{F}{({\phi_0})_{box}}\},
\end{aligned}
\end{equation}
where $N_{pos}$ indicates the number of positive samples, and $p_{i,j}^*$ denotes the category probability at target location $(i,j)$. $\mathbbm{1}$  represents the indicator function, which ensures only the positive instance mask samples perform the level set evolution.  $\mathbbm{1}$ is set to one if $p_{i,j}^* > 0 $, and zero otherwise.  $\alpha$ is the weight parameter, which is set to $3.0$ empirically in our implementation. 

\noindent \textbf{Inference.} It is worth noting that the level set evolution is only employed during training to generate implicit supervisions for network optimization. \textit{The inference process is the same as the original SOLOv2 network.} Given the input image, the mask prediction is directly generated with efficient matrix non-maximum suppression (NMS).
Comparing to SOLOv2, our proposed network introduces only one additional convolution layer to generate the high-level features with negligible cost.

\section{Experiments}
To evaluate our proposed approach, we conduct experiments on four challenging datasets, including Pascal VOC~\cite{pascalvoc2010} and COCO~\cite{lin2014microsoft}, remote sensing dataset iSAID~\cite{cvpr2019isaid} and medical dataset LiTS~\cite{bilic2019lits}. On all datasets, \textit{only box annotations are used during training.} 

\begin{table*}[bt]
	\centering
	\caption{\textbf{Performance comparisons} on  Pascal VOC \texttt{val} 2012. ``$*$" denotes the results of GrabCut reported from BoxInst~\cite{cvpr2021_boxinst}. All entries are the results using \textit{box-supervision}.
	}
	\begin{tabular}{ l | l |ccccc}
		~~methods & backbone & AP &AP$_{25}$ & AP$_{50}$ & AP$_{70}$& AP$_{75}$~ \\
		\Xhline{1pt}
		~~GrabCut$^*$~\cite{TOG2004grabcut} & ResNet-101 & 19.0 & - & 38.8 & - & 17.0~ \\
		~~SDI~\cite{cvpr2017SDI} & VGG-16 & - & - & 44.8 & - & 16.3~ \\
		~~Liao \textit{et al.}~\cite{liao2019weakly} & ResNet-101 & - & - & 51.3 & - & 22.4~ \\
		~~Sun \textit{et al.}~\cite{sun2020weakly} & ResNet-50 & - &- & 56.9 & - & 21.4~ \\ 
		~~BBTP~\cite{nips2019-bbtp} & ResNet-101 & 23.1 & - & 54.1 & - & 17.1~ \\
		~~BBTP w/ CRF~\cite{nips2019-bbtp} & ResNet-101 & 27.5 & - &59.1 & - &21.9~ \\
		~~Arun \textit{et al}.~\cite{arun2020weakly} & ResNet-101 & - & 73.1  &  57.7 & 33.5 & 31.2~ \\
		~~BBAM~\cite{cvpr2021bbam} & ResNet-101 & - & 76.8 & 63.7  & 39.5 & 31.8~ \\
		~~BoxInst~\cite{cvpr2021_boxinst} & ResNet-50 & 34.3 &- & 59.1 & -  & 34.2~\\ 
		~~BoxInst~\cite{cvpr2021_boxinst} & ResNet-101 & 36.5 &- &  61.4 & - & 37.0~\\ 
		~~DiscoBox~\cite{iccv2021discobox} & ResNet-50 &- &71.4 & 59.8 & 41.7 & 35.5~ \\
		~~DiscoBox~\cite{iccv2021discobox}& ResNet-101 &- & 72.8 & 62.2 & 45.5 & 37.5~\\
		\hline
		
		~~\textbf{Ours} & ResNet-50 & 36.3 &  76.3 & 64.2  & 43.9 & 35.9~  \\
		~~\textbf{Ours} & ResNet-101 & \textbf{38.3} & \textbf{77.9} & \textbf{66.3} & \textbf{46.4} & \textbf{38.7}~  \\
		
	\end{tabular}
	
	\label{tab:vocsota}
\end{table*}

\begin{table*}[bt]
	\centering
	\caption{\textbf{Instance segmentation mask AP} (\%)  on the COCO \texttt{test}-\texttt{dev}. ``$\dag$" denotes the result of BBTP on the COCO \texttt{val2017} split. ``$*$" indicates that the BoxCaseg is trained with box and salient object supervisions.} 
	\begin{tabular}{ l | l |cccccc}
		~~~~method  & backbone &AP & AP$_{50}$ & AP$_{75}$&AP$_{S}$ & AP$_{M}$ & AP$_{L}$  \\
		\Xhline{1pt}
		
		\emph{mask-supervised:} &&&&&&&\\
		~~~~Mask R-CNN~\cite{iccv2017maskrcnn}                    &ResNet-101 &35.7  &58.0   &37.8   &15.5   &38.1   &52.4 \\
		
		~~~~YOLACT-700~\cite{iccv2019yolact} & ResNet-101 &  31.2 &50.6& 32.8& 12.1& 33.3 &47.1   \\
		
		
		~~~~PolarMask~\cite{cvpr_2020polarmask} & ResNet-101  & 32.1&53.7 &33.1& 14.7 &33.8 & 45.3 ~ \\
		
		~~~~CondInst~\cite{eccv2020_condinst} & ResNet-101  &  39.1 &60.9 & 42.0 & 21.5 & 41.7 & 50.9~ \\
		
		
		~~~~SOLOv2~\cite{wang2020solov2} &ResNet-101 & 39.7 &60.7 &42.9& 17.3& 42.9& 57.4~\\
		\hline\hline
		
		\hline
		\emph{box-supervised:} &&&&&&&\\
		~~~~BBTP$^\dag$~\cite{nips2019-bbtp} & ResNet-101 & 21.1 &45.5 &17.2& 11.2 &22.0& 29.8~ \\
		
		~~~~BBAM~\cite{cvpr2021bbam} & ResNet-101 & 25.7 & 50.0& 23.3 & - & -& -~ \\ 
		
		~~~~BoxCaseg*~\cite{cvpr2021boxcaseg} & ResNet-101  &30.9 & 54.3 & 30.8 & 12.1 & 32.8 & 46.3~ \\
		
		
		~~~~BoxInst~\cite{cvpr2021_boxinst} & ResNet-101 & 33.2 & 56.5& 33.6& 16.2 & 35.3& 45.1~ \\ 
		~~~~BoxInst~\cite{cvpr2021_boxinst} & ResNet-101-DCN & 35.0 & \textbf{59.3} & 35.6& \textbf{17.1}& 37.2& 48.9~ \\ 
		
		~~~~\textbf{Ours} & ResNet-101 & 33.4 & 56.8 & 34.1 & 15.2 & 36.8 & 46.8~ \\
		~~~~\textbf{Ours} & ResNet-101-DCN & \textbf{35.4} & 59.1 & \textbf{36.7} & 16.8 & \textbf{38.5} & \textbf{51.3}~ \\
	\end{tabular}
	\label{tab:coco_results}
\end{table*}

\begin{table}[t]
	\centering
	\caption{\textbf{Deep variational instance segmentation methods} on COCO \texttt{val}. ``Sup." denotes the form of supervision, i.e., \textit{Mask} or \textit{Box}. \textit{Our method is only supervised with box annotations yet achieves competitive results}.}
	\begin{tabular}{ l | l |c|c}
		~~method & backbone & Sup. & AP~  \\
		\Xhline{1pt}
		~~DeepSnake~\cite{peng2020deepsnake}  & DLA-34~\cite{yu2018dla34} & \textit{Mask} &30.5~\\
		~~Levelset R-CNN~\cite{eccv2020levelset} &ResNet-50 & \textit{Mask} & ~34.3~ \\
		~~DVIS-700~\cite{yuan2020deep} & ResNet-50 & \textit{Mask} &~32.6~  \\
		~~DVIS-700~\cite{yuan2020deep} & ResNet-101 & \textit{Mask} & ~\textbf{35.7}~   \\
		\hline
		~~\textbf{Ours} & ResNet-101 & \textbf{\textit{Box}} & ~33.0~ \\
		~~\textbf{Ours} & ResNet-101-DCN & \textbf{\textit{Box}} & ~35.0~ \\	
	\end{tabular}
	\label{tab:deepvaria}
\end{table}
\subsection{Datasets}
\textbf{Pascal VOC}~\cite{pascalvoc2010}. Pascal VOC consists of 20 categories. As in~\cite{nips2019-bbtp, cvpr2021_boxinst,cvpr2021bbam}, the augmented Pascal VOC 2012~\cite{iccv2011_SBDdataset} dataset is used, which contains 10, 582 images for training and 1, 449 validation images for evaluation. 

\noindent \textbf{COCO}~\cite{lin2014microsoft}. COCO has 80 general object classes. Our models are trained on \texttt{train2017} (115K images), and evaluated on \texttt{val2017} (5K images) and \texttt{test-dev} split (20K images).

\noindent \textbf{iSAID}~\cite{cvpr2019isaid}. It is a large-scale high-resolution remote sensing dataset for aerial instance segmentation, containing many small objects with  complex backgrounds. The dataset comprises 1,411 images for training and 458 validation images for evaluation with 655,451 instance annotations.
 
\noindent \textbf{LiTS}~\cite{bilic2019lits}. The Liver Tumor Segmentation Challenge (LiTS) dataset$\footnote{https://competitions.codalab.org/competitions/17094}$ consists of 130 volume CT scans for training and 70 volume CT scans for testing. We randomly partition all the scans having mask labels into the training and validation dataset with the ratio of 4:1.

\subsection{Implementation Details}

The models are trained with the AdamW~\cite{AdamW2017decoupled}
optimizer on 8 NVIDIA V100 GPUs. 
The training schedules of ``1$\times$" and ``3$\times$" are the same as \texttt{mmdetection} framework~\cite{chen2019mmdetection} with 12 epochs and 36 epochs, respectively.
ResNet~\cite{he2016deep} is employed as the backbone, which is initialized with the ImageNet~\cite{deng2009imagenet} pre-training weights.  
For COCO, the initial learning rate is $10^{-4}$ with 16 images per mini-bath. For Pascal VOC, the initial learning rate is $5\times 10^{-5}$ with 8 images per mini-bath.
The scale jitter is used  where the shorter image side is randomly sampled from 640 to 800 pixels on COCO and Pascal VOC datasets for fair comparison. 
For iSAID and LiTS, all the models on each dataset are trained with the same settings.
COCO-style mask AP (\%) is adopted for performance evaluation. Following~\cite{cvpr2021bbam, iccv2021discobox}, we also report the average precision (AP) at four IoU thresholds (including 0.25, 0.50, 0.70 and 0.75) for the comparison on Pascal VOC dataset. 
The non-negative weight $\gamma$ in Eq.~\ref{eq4} is set to $10^{-4}$ by default. 
\begin{table}[t]
	\centering
	\caption{\textbf{Results of mask AP} (\%) on iSAID \texttt{val}. All models are trained with ``1$\times$" schedule (12 epoch) with 600$\times$600 input size. }
	\begin{tabular}{ l | l |c|ccc}
		~~method & backbone & Sup. & AP &AP$_{50}$ & AP$_{75}$~  \\
		\Xhline{1pt}
		~~Mask R-CNN~\cite{iccv2017maskrcnn}  &R-50-C4 & \textit{Mask} & 28.8 & 51.8 & 27.7~ \\
		~~PolarMask~\cite{cvpr_2020polarmask} & R-50-FPN & \textit{Mask} &  27.2 & 48.5 & 27.3~ \\
		~~CondInst~\cite{eccv2020_condinst} & R-50-FPN & \textit{Mask} & 29.5  & 54.5 & 28.3~ \\
		\hline
		~~BoxInst~\cite{cvpr2021_boxinst} & R-50-FPN & \textit{Box} & 17.8 & 41.4 & 12.9~ \\
		~~\textbf{Ours} & R-50-FPN & \textit{Box} & \textbf{20.1} & \textbf{41.8} & \textbf{16.6}~ \\
	\end{tabular}
	\label{tab:isaid}
\end{table}

\begin{table}[t]
	\centering
	\caption{\textbf{Instance segmentation results} on LiTS \texttt{val}. All models are trained with ``1$\times$" schedule (12 epoch). Our method outperforms BoxInst by 3.8\% AP.}
	\begin{tabular}{ l | l |c|ccc}
		~~method & backbone & Sup. & AP &AP$_{50}$ & AP$_{75}$  \\
		\Xhline{1pt}
		~~Mask R-CNN~\cite{iccv2017maskrcnn}  &R-50-FPN & \textit{Mask} & 64.2 & 81.6 & 71.0   \\
		\hline
		~~BoxInst~\cite{cvpr2021_boxinst} & R-50-FPN & \textit{Box} & 40.7 & 67.8& 40.2 \\
		~~\textbf{Ours} & R-50-FPN & \textit{Box} & \textbf{44.5} & \textbf{78.6} & \textbf{45.6} \\
	\end{tabular}
	\label{tab:lits}
\end{table}

\subsection{Main Results}
We compare our proposed method against the state-of-the-art instance segmentation approaches, including box-supervised and fully mask-supervised methods in different scenarios.

Most box-supervised methods are evaluated on the Pascal VOC dataset. Table~\ref{tab:vocsota} reports the comparison results. Our method outperforms BoxInst~\cite{cvpr2021_boxinst} by 2.0\%  and 1.8\% AP with ResNet-50 and ResNet-101 backbones, respectively, achieving the best performance. For AP$_{25}$ and AP$_{50}$, our method can obtain 77.9\% and 66.3\% accuracy, largely outperforming the recent DiscoBox~\cite{iccv2021discobox} by 5.1\% and 4.1\%. The high IoU threshold-based AP metrics can reflect the segmentation performance with accurate boundary, which is in line with the practical application. Our approach achieves  38.7\% AP$_{75}$ with ResNet-101, which outperforms BoxInst~\cite{cvpr2021_boxinst} and DiscoBox~\cite{iccv2021discobox} by 1.7\% and 1.2\%, respectively.

\begin{figure}[t]
	\centering
	\includegraphics[width=1.0\linewidth]{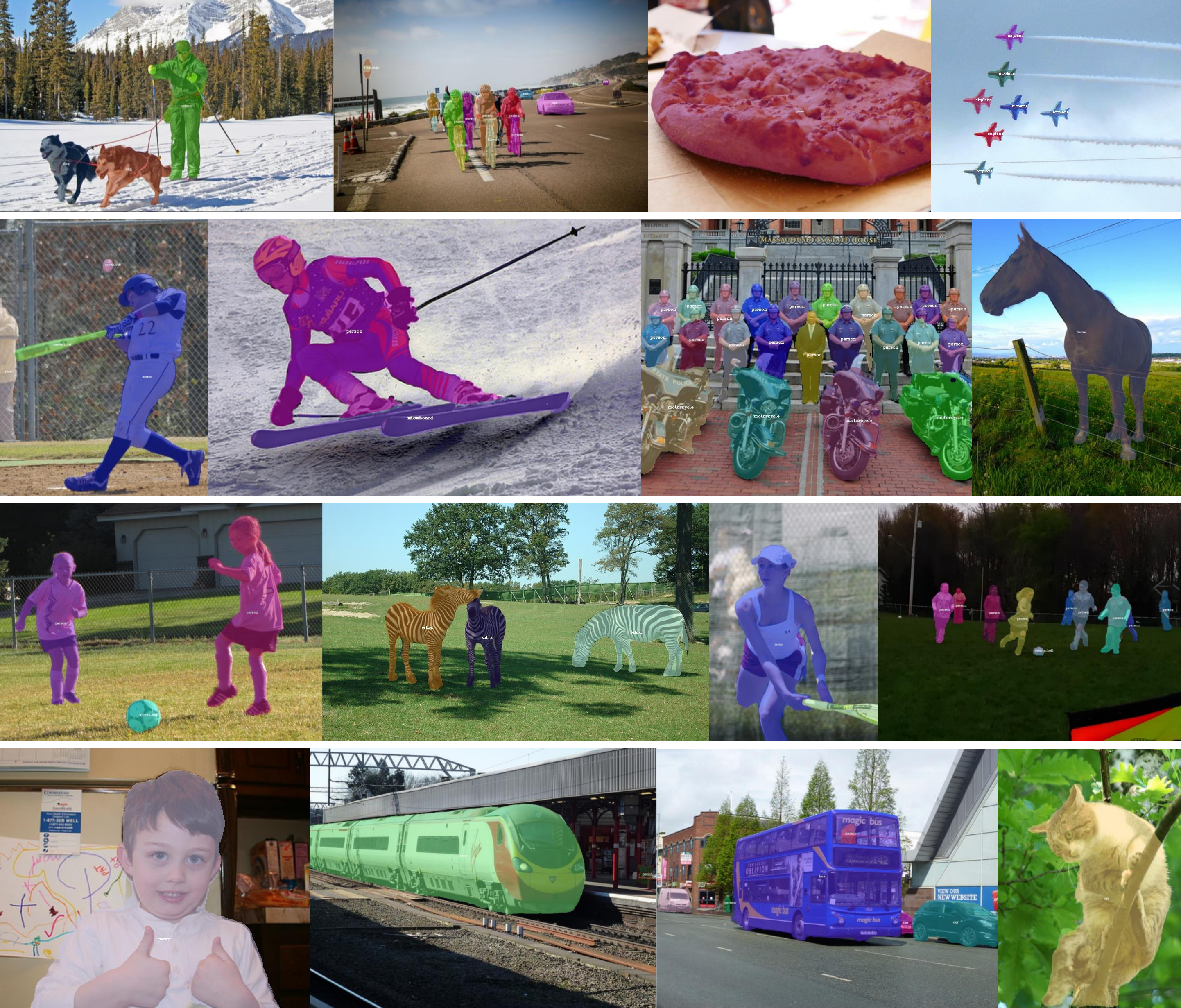}
	\caption{\textbf{Visualization of instance segmentation results} on general scene. The model is trained with only box annotations.}
	\label{fig:coco_vis}
\end{figure}

Table~\ref{tab:coco_results} shows the main results on COCO \texttt{test-dev} split.
Both fully mask-supervised and box-supervised methods are compared in the evaluation. Our method outperforms BBTP~\cite{nips2019-bbtp} by  12.3\% AP with the same backbone. In contrast to the recent box-supervised methods, our method outperforms BBAM~\cite{cvpr2021bbam} and  BoxCaseg~\cite{cvpr2021boxcaseg} by 7.7\% AP and 2.5\% AP using  ResNet-101. It achieves 33.4\% AP and 35.4\% AP, which is higher than BoxInst~\cite{cvpr2021_boxinst} by 0.2\% and 0.4\% with ResNet-101 and ResNet-101-DCN backbones, respectively. Our approach achieves  16.8\% AP$_S$ on small objects, which is slightly lower than BoxInst~\cite{cvpr2021_boxinst} by 0.3\%. This is because small objects lack rich features for level set evolution to distinguish the foreground object and background within the bounding box. However, our method obtains the best results for large objects, largely outperforming BoxInst~\cite{cvpr2021_boxinst} by 2.4\% AP$_L$ using the same ResNet-101-DCN. Our method even performs better than some recent fully mask-supervised methods, such as YOLACT~\cite{iccv2019yolact} and PolarMask~\cite{cvpr_2020polarmask}. This shows that our method narrows the performance gap between mask-supervised and box-supervised instance segmentation.
Fig.~\ref{fig:coco_vis} visualizes some instance segmentation results on COCO and  Pascal VOC datasets.

We then compare our method with other deep variational-based instance segmentation approaches. DeepSnake~\cite{peng2020deepsnake} is based on the classical snake method~\cite{kass1988snakes}. Levelset R-CNN~\cite{eccv2020levelset} and DVIS-700~\cite{yuan2020deep} are also built on level set function. These methods are all fully supervised by the mask annotations. 
As shown in Table~\ref{tab:deepvaria}, our method achieves comparable results to the fully supervised variational-based methods, and even outperforms DeepSnake~\cite{peng2020deepsnake} and Levelset R-CNN~\cite{eccv2020levelset}.

\begin{figure*}[t]
	\centering
	\includegraphics[width=0.975\linewidth]{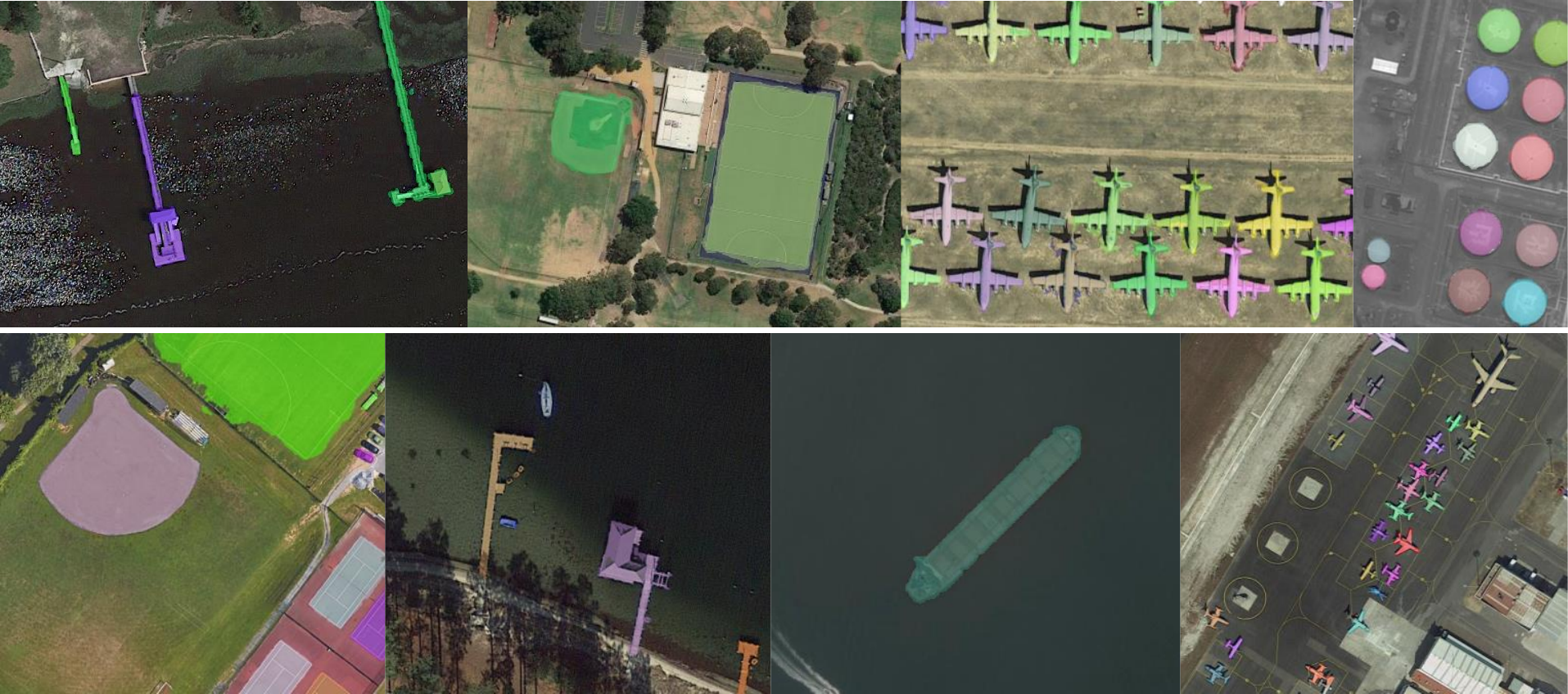}
	\caption{\textbf{Visual results } of iSAID \texttt{val}. The mask predictions are obtained on the high-resolution remote sensing images only with box supervision.} 
	\label{fig:isaid_vis}
\end{figure*}

\begin{figure*}[t]
	\centering
	\includegraphics[width=0.975\linewidth]{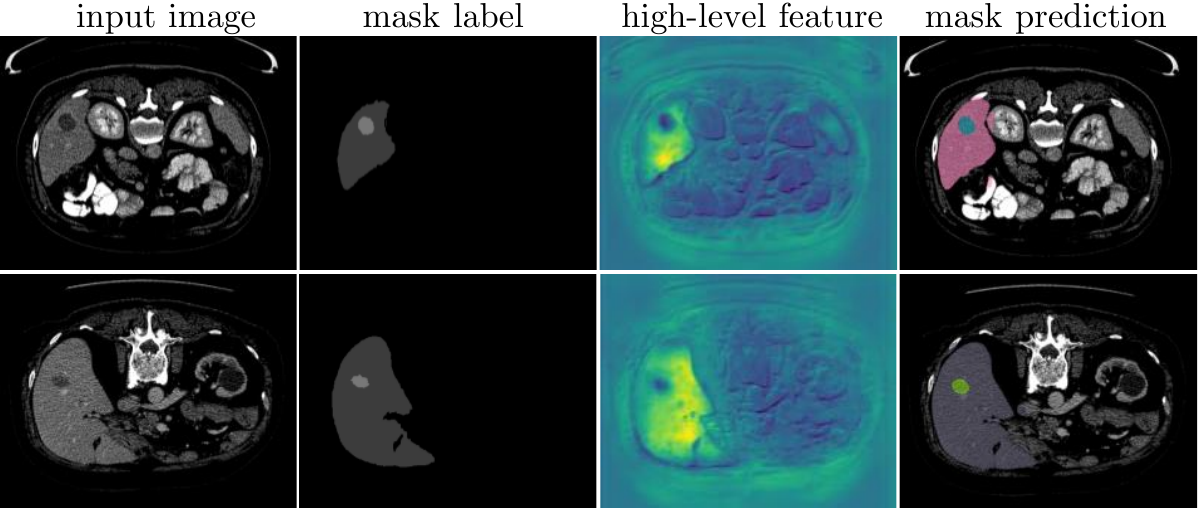}
	\caption{\textbf{Visualization examples} of LiTS \texttt{val}. The high-level feature represents the input deep feature for level set evolution.} 
	\label{fig:lits_vis}
\end{figure*}

To further validate the robust performance of our method in more complicated scenarios, 
we conduct experiments on remote sensing and medical image datasets. In remote sensing, the objects of the same class are densely-distributed. For medical images, the background is highly similar to the foreground. The previous pixel relationship model-based methods are built on the neighbouring pixel pairs. They are easily affected by the noisy context. Our level set-based method drives the curve to fit the object boundary under the guidance of level set minimization, which is more robust.
Table~\ref{tab:isaid} and Table~\ref{tab:lits} show the mask AP results on iSAID and LiTS datasets, respectively. It can be clearly seen that our approach outperforms BoxInst~\cite{cvpr2021_boxinst} by 2.3\% AP on iSAID and 3.8\% AP on LiTS. Fig.~\ref{fig:isaid_vis} and 
Fig.~\ref{fig:lits_vis} show several examples of instance segmentation on iSAID and LiTS, respectively. One can see that our method is effective in various scenarios.

\subsection{Ablation Experiments}
The ablation study is conducted on Pascal VOC dataset to examine the effectiveness of each module in our proposed framework.
\begin{table}[t]
	\centering
	\caption{The impact of \textbf{level set energy} with different settings. $I_{u}$ and $I_{f}$ denote the input image and high-level feature as the input data terms of energy, respectively. $\mathcal{B}$ and $I$ represent the $\Omega$ space  of bounding box or the full-image region for level set evolution.}
	\begin{tabular}{ c c  c |c c|ccc}
		~$\mathcal{F}_{\phi_0}$ & ~$\mathcal{F_{\phi}}({I_{u}})$ & ~$\mathcal{F}_{\phi}{(I_{f})}$ & ~$\Omega \in \mathcal{B}$ & ~$\Omega \in I$ &  AP &AP$_{50}$ & AP$_{75}$~\\
		\Xhline{1pt}
		~$\checkmark$ &  & & ~$\checkmark$& &19.7 & 47.4 & 13.9  \\ 
		~$\checkmark$ &~$\checkmark$ & & ~$\checkmark$& & 22.2 & 49.5 & 17.4 \\
		~$\checkmark$ &~$\checkmark$ &~$\checkmark$ &~$\checkmark$ & & \textbf{24.7} & \textbf{53.3} & \textbf{20.8} \\
		~$\checkmark$ &~$\checkmark$ &~$\checkmark$ & &~$\checkmark$ & 21.7 & 48.4 & 17.4 \\
	\end{tabular}
	\label{tab:lossterm}
	\vspace{-0.1in}
\end{table}

\noindent \textbf{Level Set Energy.} We firstly investigate the impact of level set energy functional with different settings. Table~\ref{tab:lossterm} gives the evaluation results. Our method achieves 19.7\% AP only with the box projection function as $\mathcal{F}_{\phi_0}$ to drive the network to initialize the boundary during training. This indicates that the initialization for level set function is effective to generate the initial boundary. When the original image $I_{u}$ is employed as the input data term in Eq.~\ref{eq8}, our method can achieve 22.2\% AP. On the other hand, our method achieves better performance with 24.7\% AP when the deep high-level features $I_{f}$ are employed as the extra input data. This demonstrates that both original image and high-level features can provide useful information for robust level set evolution. Besides, the above results are constrained within the bounding box $\mathcal{B}$ region for curve evolution. When the global region with the full-image size is regarded as the $\Omega$, there is a noticeable performance drop (24.7\% vs. 21.7\%). This indicates that the bounding box region can make the level set evolution smoother with less noise interference.

\noindent \textbf{Number of Channels for High-level Feature.} Secondly, we investigate the selection of the total number of channels for the output high-level feature $I_{f}$. As shown in Table~\ref{tab:channelnum}, our method obtains better representation with 24.7\% AP performance when the number of channels $C_{I_f}$ is set to 9. When $C_{I_f} = 10$, the performance drops (24.7\% vs. 22.0\%). This indicates that the more channels may introduce uncertain semantic information for level set evolution.

\begin{table}[t]
	\footnotesize
	\centering
	\begin{minipage}[t]{0.30\linewidth}
		\centering
		\caption{Different channel number $C_{I_f}$ of high-level features for curve evolution.} 
		\begin{tabular}{c|ccc}
			$C_{I_f}$& AP &AP$_{50}$ & AP$_{75}$\\
			\Xhline{1pt}
			5 & 23.3 & 51.3 & 18.7 \\
			8 & 24.4 & 52.1 & 20.1 \\
			9 & \textbf{24.7} & \textbf{53.3} &  \textbf{20.8}\\
			10 & 22.0 & 49.2 & 17.3 \\
			11 & 21.9 & 49.4 & 16.9 \\			
		\end{tabular}
		\label{tab:channelnum}
	\end{minipage}
	\hspace{0.10cm}
		\begin{minipage}[t]{0.30\linewidth}
		\centering
		\caption{Training schedules with ``1$\times$" single-scale training and ``3$\times$" multi-scale training.} 
		\begin{tabular}{ c|ccc}
			sched.&  AP &AP$_{50}$ & AP$_{75}$\\
			\Xhline{1pt}
			1$\times$  & 24.7 & 53.3 & 20.8 \\
			3$\times$ &\textbf{34.4} & \textbf{62.2.} & \textbf{34.6} \\
		\end{tabular}
		\label{tab:trainingschdule}
	\end{minipage}
	\hspace{0.10cm}
	\begin{minipage}[t]{0.30\linewidth}
		\centering
		\caption{The effectiveness of tree filter~\cite{nips2019learnable} for  high-level structural features in level set.} 
		\begin{tabular}{ c|ccc}
			tree filter  &  AP &AP$_{50}$ & AP$_{75}$\\
			\Xhline{1pt}
			w/o.  & 34.4 & 62.2 & 34.6 \\
			w.  &\textbf{36.3} & \textbf{64.2.} & \textbf{35.9} \\
		\end{tabular}
		\label{tab:treefilter}
	\end{minipage}
	\label{tab:ablation}
\end{table}

\noindent \textbf{Training Schedule.} We evaluate the proposed network using different training schedules. Table~\ref{tab:trainingschdule} shows the results with 12 epochs (1$\times$) and 36 epochs (3$\times$). It can be observed that a longer training schedule benefits the performance of our method. Due to the relatively small size of Pascal VOC compared with COCO (about $1/10$), longer training schedule leads to significant improvement (24.7\% vs. 34.4\%). This implies that level set evolution needs more training time to achieve better convergence for instance segmentation. 

\noindent \textbf{Effectiveness of Deep Structural Feature.} 
We study the impact of tree filter~\cite{nips2019learnable}, which models long-range dependencies and preserves object structure, on obtaining deep semantic features for level set evolution. Table~\ref{tab:treefilter} shows the results. One can see that by applying the tree filter to high-level deep features, +1.9\% AP improvement can be achieved. 

\section{Conclusion}
This paper presented a single-shot box-supervised instance segmentation approach that iteratively learns a series of level set functions in an end-to-end fashion. An instance-aware mask map was predicted and used as the level set, and both the original image and deep high-level features were employed as the inputs to evolve the level set curves, where a box projection function was employed to obtain the initial boundary. By minimizing the fully differentiable energy function, the level set for each instance was iteratively optimized within its corresponding bounding box annotation. Extensive experiments were conducted on four challenging benchmarks, and our proposed approach demonstrated leading performance in various scenarios. 
Our work narrows the performance gap between fully mask-supervised and box-supervised instance segmentation.

\section*{Acknowledgments}
This work is supported by National Natural Science Foundation of China under Grants (61831015) and Alibaba-Zhejiang University Joint Institute of Frontier Technologies.

%
%
\bibliographystyle{splncs04}
\bibliography{egbib}
\end{document}